\documentclass{Interspeech2024}

\usepackage{todonotes}
\usepackage{multirow}
\usepackage{graphicx}
\usepackage{float}
\usepackage{soul}

\usepackage{inconsolata}
\usepackage[activate={true,nocompatibility},kerning=true,spacing=true, stretch=10,shrink=30]{microtype}
\usepackage{listings}
\usepackage{textcomp}
\microtypecontext{spacing=nonfrench}

\interspeechcameraready 
\def\canary{Canary}

\title{Less is More: Accurate Speech Recognition \& Translation\\ without Web-Scale Data}

\name{Krishna C.}{Puvvada$^*$}
\name{Piotr}{Żelasko$^*$}
\name{He}{Huang$^*$}
\name{Oleksii}{Hrinchuk}
\name{Nithin Rao}{Koluguri}
\name{Kunal}{Dhawan}
\name{Somshubra}{Majumdar}
\name{Elena}{Rastorgueva}
\name{Zhehuai}{Chen}
\name{Vitaly}{Lavrukhin}
\name{Jagadeesh}{Balam}
\name{Boris}{Ginsburg}

\address{NVIDIA, Santa Clara, CA, USA}

\email{\{kpuvvada,pzelasko,heh\}@nvidia.com}


\begin{document}
\maketitle

\begin{abstract}    
Recent advances in  speech recognition and  translation rely on hundreds of thousands of hours of Internet speech data. 
We argue that state-of-the art accuracy can be reached without relying on web-scale data.   \emph{\canary{}} - multilingual ASR and speech translation model, 
outperforms current state-of-the-art models -- Whisper, OWSM, and Seamless-M4T on  English, French, Spanish, and German  languages, while being trained on an order of magnitude less data than these models. Three key factors  enables such data-efficient model: (1) a FastConformer-based attention encoder-decoder architecture (2) training on synthetic data generated with machine translation and (3) advanced training techniques: data-balancing, dynamic data blending, dynamic bucketing and noise-robust fine-tuning. The model, weights, and training code will be open-sourced.
\end{abstract}

\renewcommand{\thefootnote}{\fnsymbol{footnote}}
\footnotetext[1]{Equal contribution}
\renewcommand{\thefootnote}{\arabic{footnote}}

\section{Introduction}
\label{sec:intro}

The landscape of automatic speech recognition (ASR) and automatic speech translation (AST) has been disrupted with the introduction of large scale multi-task models.

Whisper~\cite{radford2022robust} is a transformer~\cite{vaswani2017attention} attention encoder-decoder (AED) model~\cite{bahdanau2015neural} that has demonstrated impressive ASR and AST capabilities in 96 languages. It was trained initially with 680K hours of data and later extended to 5M hours, out of which 4M were transcribed by an earlier model version.

Seamless~\cite{barrault2023seamless} is a multimodal streaming translation model supporting around 100 languages. It uses several components pretrained on over 4M unlabeled hours of speech, which are later fine-tuned jointly on 125k hours.

OWSM~\cite{peng2023reproducing} is the first fully open-source attempt at reproducing  Whisper model. It's trained on 180k hours of publicly available data and supports 151 languages. The latest OWSM ver 3.1 adopted E-Branchformer architecture, achieving superior accuracy and speed~\cite{peng2024owsm}.

Beyond impressive performance, transformer architecture, and size, what these models share in common are significant resource requirements: the amount of data and time required to train them. OWSM is the only work mentioned here that informs about training time and resources used: 16 days of training on 64 NVIDIA A100 40GB GPUs. It's also trained on the least amount of data amongst models under discussion, hence we expect Whisper and Seamless would require even more resources and/or time. Another observation worth noting is that OWSM and Whisper train with a global batch size of 256 (increased to 1024 for Whisper v2 and v3), where each utterance is padded to be exactly 30s long. It was observed in~\cite{peng2024owsm} that such long utterances harm the model convergence. We also note that this approach may lead to significant padding in mini-batches, resulting in wasted computation on non-informative frames. We present an alternative approach to sampling and batching that allows us to iterate through data twice as fast, while balancing different languages and data sources better. We further accelerate the training and inference by adopting a FastConformer~\cite{rekesh2023fast} architecture and initializing the encoder from a ASR only pretrained checkpoint.

Besides ASR, all models under discussion are capable of AST, i.e., they can transcribe the recording in any of the supported languages (except for Whisper, which only translates \texttt{X$\rightarrow$en}). There exists much less data for AST than for ASR, and creating such datasets usually involves translating the transcript into the target language. We demonstrate it is possible to train a strong AST model without using existing AST data at all: instead, we adopt a text machine translation model to automatically create AST supervisions for training from ASR data.

\noindent Key contributions of this work:
\begin{itemize}
    \item We introduce \canary{}, an open-source AED model capable of ASR and AST in English, French, Spanish, and German. \canary{} outperforms other multi-task AED models of similar size on established benchmarks.
    \item We demonstrate that it is possible to train highly accurate speech translation models using only pseudo-labeled translation data.
    \item We train \canary{} under 48 hours using 128 NVIDIA A100 80GB GPUs with a total of 225K weight updates by initializing the encoder from pre-trained weights and using dynamic bucketing batching techniques.
\end{itemize}
Using all techniques mentioned above, we train \canary{} using ``only" 86K hours of speech. This is an order of magnitude less than amount of data used by Whisper and Seamless models.

\section{Methods}
\label{sec:methods}

\textbf{Model architecture.} \canary{} uses FastConformer encoder \cite{rekesh2023fast} and a Transformer decoder. FastConformer is a speech-specific modification of a transformer based on Conformer~\cite{gulati2020conformer} that increases the downsampling factor to 8, achieving 2.8x speedup without loss of modeling capacity~\cite{rekesh2023fast}. We train \canary{} with a cross-entropy objective.

\noindent \textbf{Multi-task training and prompting.} To achieve multi-task support, we condition the model's predictions using prompts. Similarly to Whisper, we adopt the following special prompt tokens: \texttt{<|startoftranscript|>}, \texttt{<|transcribe|>}, \texttt{<|translate|>}, \texttt{<|nospeech|>}, \texttt{<|endoftranscript|>}, and an additional special token for each supported language. Our prompt is constructed similarly to Whisper's, except we specify both input (audio) and output (text) languages as tokens before and after \texttt{<|translate|>}, respectively. We also add new special tokens \texttt{<|pnc|>} and \texttt{<|nopnc|>} at the end of the prompt control sequence to select whether the model should predict punctuation and capitalization, or not. We adopt SentencePiece~\cite{kudo2018sentencepiece} and concatenated tokenizer~\cite{dhawan2023unified} with a vocabulary size of 1024 for each supported language and a 32-unit sub-tokenizer for the special tokens.

\noindent \textbf{Dataset and language blending.} Since the training data consists of multiple languages and diverse datasets, we aim to ensure consistent sampling of each throughout the training process.  Failing to do so tends to result in training intervals where the model performs better on some domains/languages than others. Given $N$ datasets, we define a probability distribution $P(d)$ of choosing the next example from a specific dataset $d$. To ensure this distribution remains stationary throughout training, we use a stochastic weighted multiplexing mechanism to combine the datasets. When constructing a mini-batch of samples, for each training example we first select the dataset according to $P(d)$ and then sample an utterance from this dataset. In the expectation, each mini-batch would have a ratio of data coming from each source according to $P(d)$.\footnote{\texttt{CutSet.mux} method at https://lhotse.readthedocs.io/} We consider weights as ``natural" when they are proportional to the cumulative duration of each dataset. We also experiment with re-weighting strategies based on stratification by language and, within each language, by dataset. Additionally, we explore the use of temperature scaling applied to these dataset weights before sampling.\footnote{Note that this approach is compatible with optimized dataloading techniques that rely on sequential I/O (such as webdataset\cite{webdataset} or Lhotse Shar\cite{lhotseshar}) when each dataset is stored separately (e.g., as a separate collection of tar file shards).}

\noindent \textbf{Variable utterance length and dynamic batch sizes.} We address the issue of duration distribution variability across utterances, typically ranging between one to forty seconds. We perform stratified sampling based on sample duration with a technique known as bucketing, where there are $M$ buckets, each containing utterances of similar duration, and any given mini-batch is sampled from just one bucket chosen randomly. Unlike most bucketing implementations that require partitioning data up-front, we leverage a dynamic bucketing technique. We estimate the optimal bucket bins (in the sense of equal bucket utilization given data duration distribution) up-front, but the allocation of utterances into buckets is done dynamically during training with a small utterance buffer in CPU RAM. The mini-batches are sampled to satisfy a maximum total duration, resulting in dynamic batch size.\footnote{\texttt{DynamicBucketingSampler} at https://lhotse.readthedocs.io/} We further account for quadratic sequence length complexity of the encoder by introducing a quadratic duration penalty. We find it helps equalize the GPU utilization across mini-batches from different buckets and improves the throughput. Thanks to this data-driven bucketing calibration, we typically observe only about 3\% of padding in mini-batches compared to as much as 50\% when using non-stratified sampling.

\noindent \textbf{Improving robustness to hallucinations.} In this work, we define hallucinations as producing transcriptions when input audio contains no speech. The severity of hallucinations is influenced by both the model architecture and the loss function employed. The tendency to hallucinate is a known flaw of AED models. We find that adding noise, music, and other non-speech data to the training dataset as a pseudo-language with its own weight significantly reduces hallucinations, though doesn't eliminate them entirely.

\section{Experimental setup}
\label{sec:exp}

\textbf{Training data.} \canary{} was trained on a mixture of public and in-house datasets. Table~\ref{tab:asr_data} shows the composition of training data for the ASR task (81.5K hours in total). 
The public portion of English is composed of LibriSpeech, Fisher Corpus, Switchboard-1, WSJ-0 \& WSJ-1, National Speech Corpus (Part 1, Part 6), VCTK, VoxPopuli (EN), Europarl-ASR (EN), Multilingual LibriSpeech (MLS)-EN  (2k hour subset), Mozilla Common Voice (MCV)-7.0, People's Speech (12K hour subset), MCV-11.0 (1.5k hour subset). 
800 hour subset of MCV-12.0, 1.5K hour subset of MLS and 200 hour subset of VoxPopuli were gathered from public sources for German. 
For Spanish, 395 hours from MCV-12.0, 780 hours from MLS, 108 hours from VoxPopuli and 141 hours from Fisher were collected from public domain. 
Similarly for French, 708 hour subset from MCV-12.0, 926 hours from MLS and 165 hours from VoxPopuli were used from public domain. Table~\ref{tab:asr_data} also shows number of hours with punctuation and capitalization (PnC) for each language. PnC was obtained from respective dataset metadata when available (e.g. LibriSpeech). Alternatively, PnC was restored using neural models for some datasets. PnC data was further processed to remove all punctuation marks except five (',?.!). Text normalization was applied to ground truth. 300 hours of non-Speech data (AudioSet strongly-labelled subset portion of \cite{kim-etal-2019-audiocaps}) is used to improve model robustness. Data for AST was solely obtained by generating synthetic labels using Neural Machine Translation models \cite{MegatronNMTEnAny2024, MegatronNMTAnyEn2024} without using additional datasets. 43K hours of English ASR data from Table~\ref{tab:asr_data} was used to generate training data for English$\rightarrow$X (X: German, Spanish, French). All available data from Table~\ref{tab:asr_data} for German, Spanish, French languages was used to prepare X $\rightarrow$ English direction of translation data. Further, a 4.8K hour English $\rightarrow$ German translation dataset from \cite{hrinchuk-etal-2023-nvidia} was also used, which in-itself was also pseudo-labeled, bringing the total size to 86.3K hours. 

\noindent \textbf{Test data.} Test sets from MCV-16.1, MLS, and VoxPopuli were used to measure ASR performance across all languages. The translation capabilities of the models were examined using FLEURS, mExpresso, and CoVoST-v2. Additionally, standard English test sets such as AMI, Earnings22, Gigaspeech, LibriSpeech (test-clean \& test-other), SPGI, and Tedlium were utilized to evaluate model generalization across different domains. The Music and Noise subsets (a total of 48 hours) from MUSAN~\cite{musan2015} were used to assess robustness to hallucinations.

\begin{table}[t]
\caption{Training data statistics with breakdown per language and availability of punctuation and capitalization (PnC).}
\label{tab:asr_data}
\resizebox{\columnwidth}{!}{%
\begin{tabular}{l|c|c|c|c}
\hline
\multicolumn{1}{c|}{\textbf{Language}} & \textbf{\begin{tabular}[c]{@{}c@{}}Public \\ + In-house [K hours]\end{tabular}} & \begin{tabular}[c]{@{}c@{}}\textbf{PnC} \\  \textbf{[K hours]}\end{tabular} & \begin{tabular}[c]{@{}c@{}}\textbf{Dur. [s]} \\  \textbf{[Min, Max]}\end{tabular}  & \begin{tabular}[c]{@{}c@{}}\textbf{\# Utterances} \\  \textbf{[M]}\end{tabular} \\ \hline \hline
English                                     & 25.5 + 37.9                                                               & 38.5              & [1,  40]                                        & 24                                  \\ \hline
German                                     & 2.5 + 3.6                                                                 & 6.1               & [1,  20]                                        & 2.4                                  \\ \hline
Spanish                                     & 1.4 + 5.2                                                                 & 1.4               & [1, 20]                                         & 3.8                                  \\ \hline
French                                     & 1.8 + 3.3                                                                 & 1.8               & [0.5, 40]                                        & 2.5                                  \\ \hline
Non-speech                             & 0.3 + 0.0                                                                  & NA                 & [0.47, 10]                                       & 0.1                                 \\ \hline \hline
Total                                  & 31.5 + 50                                                                 & 47.8              & NA                                        & 32.8                                 \\ \hline
\end{tabular}%
}
\end{table}

\begin{table*}[t]
\centering
\caption{WER (\%) results on ASR benchmarks. Baseline model predictions are generated using respective public checkpoints. Ground-truth and predictions are normalized using WhisperNormalizer\cite{radford2022robust}. \canary{} achieves lowest WER in 10 out of 12 test sets across all languages.}
\label{tab:asr_wer}
\resizebox{\textwidth}{!}{
\begin{tabular}{l|rrr|rrr|rrr|rrr}
\hline
\multicolumn{1}{c|}{\multirow{2}{*}{Model (WER $\downarrow$)}} & \multicolumn{3}{c|}{En} & \multicolumn{3}{c|}{De} & \multicolumn{3}{c|}{Es} & \multicolumn{3}{c}{Fr} \\ 
\multicolumn{1}{c|}{} & \multicolumn{1}{l|}{MCV-16.1} & \multicolumn{1}{l|}{MLS} & \multicolumn{1}{l|}{VoxPopuli} & \multicolumn{1}{l|}{MCV-16.1} & \multicolumn{1}{l|}{MLS} & \multicolumn{1}{l|}{VoxPopuli} & \multicolumn{1}{l|}{MCV-16.1} & \multicolumn{1}{l|}{MLS} & \multicolumn{1}{l|}{VoxPopuli} & \multicolumn{1}{l|}{MCV-16.1} & \multicolumn{1}{l|}{MLS} & \multicolumn{1}{l}{VoxPopuli} \\ \hline \hline
OWSM-v3.1 (1.02B) & \multicolumn{1}{r|}{11.87} & \multicolumn{1}{r|}{5.37} & 7.04 & \multicolumn{1}{r|}{9.24} & \multicolumn{1}{r|}{10.49} & 16.25 & \multicolumn{1}{r|}{9.59} & \multicolumn{1}{r|}{8.84} & 10.17 & \multicolumn{1}{r|}{13.69} & \multicolumn{1}{r|}{11.75} & 12.61 \\ \hline
SeamlessM4T-medium (1.2B) & \multicolumn{1}{r|}{10.25} & \multicolumn{1}{r|}{7.05} & 6.06 & \multicolumn{1}{r|}{9.32} & \multicolumn{1}{r|}{8.12} & 12.95 & \multicolumn{1}{r|}{7.25} & \multicolumn{1}{r|}{5.25} & 7.31 & \multicolumn{1}{r|}{11.07} & \multicolumn{1}{r|}{7.32} & 8.77 \\ \hline
SeamlessM4T-large-v2 (2.3B) & \multicolumn{1}{r|}{\textbf{7.47}} & \multicolumn{1}{r|}{4.14} & 4.68 & \multicolumn{1}{r|}{5.82} & \multicolumn{1}{r|}{6.08} & \textbf{10.68} & \multicolumn{1}{r|}{4.82} & \multicolumn{1}{r|}{4.14} & 6.76 & \multicolumn{1}{r|}{7.75} & \multicolumn{1}{r|}{5.38} & 6.82 \\ \hline
Whisper-large-v3 (1.5B) & \multicolumn{1}{r|}{9.92} & \multicolumn{1}{r|}{3.53} & 6.23 & \multicolumn{1}{r|}{6.17} & \multicolumn{1}{r|}{5.83} & 16.50 & \multicolumn{1}{r|}{4.94} & \multicolumn{1}{r|}{4.42} & 8.01 & \multicolumn{1}{r|}{11.18} & \multicolumn{1}{r|}{5.38} & 7.52 \\ \hline \hline
\canary{} (1B) & \multicolumn{1}{r|}{7.83} & \multicolumn{1}{r|}{\textbf{3.03}} & \textbf{4.42} & \multicolumn{1}{r|}{\textbf{4.49}} & \multicolumn{1}{r|}{\textbf{4.09}} & 10.70 & \multicolumn{1}{r|}{\textbf{3.88}} & \multicolumn{1}{r|}{\textbf{3.12}} & \textbf{5.02} & \multicolumn{1}{r|}{\textbf{6.37}} & \multicolumn{1}{r|}{\textbf{4.06}} & \textbf{5.48} \\ \hline
\end{tabular}}
\end{table*}

\begin{figure*}[th]
    \centering
    \includegraphics[width=0.7\textwidth]{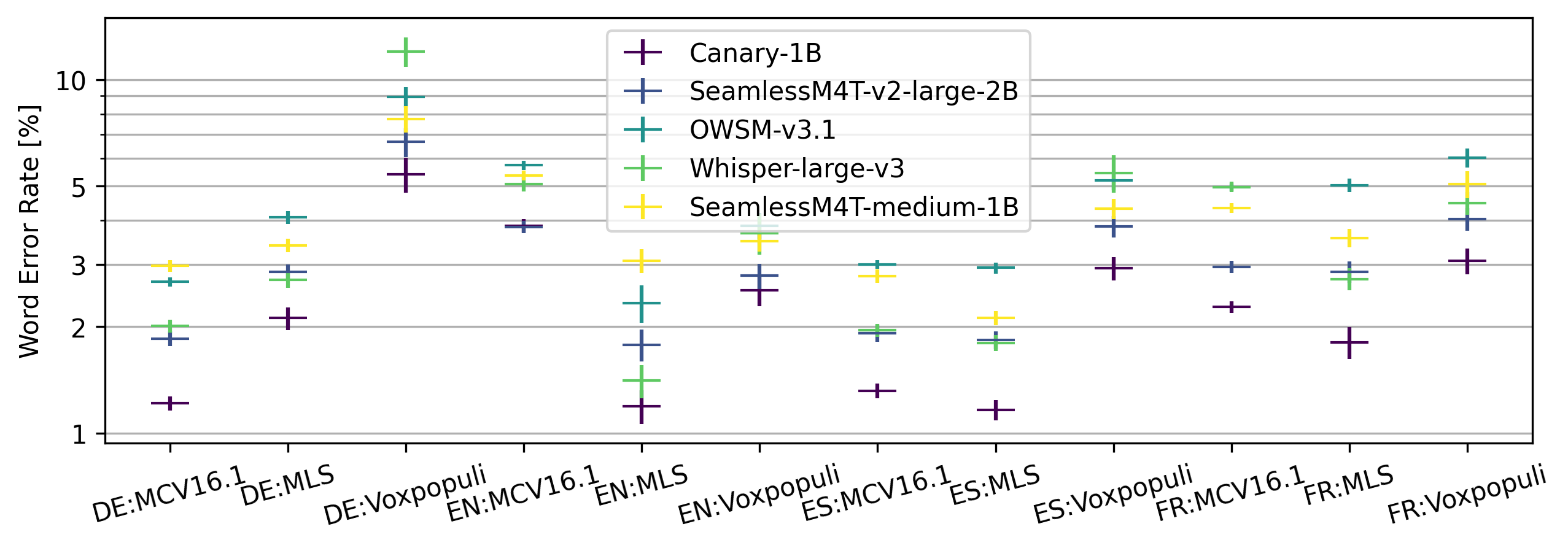}
    \caption{Word error rate on 12 test sets for the proposed model and 4 baselines. Vertical bars indicate 95\% confidence intervals obtained from boostrap method with $10^4$ replications~\cite{Bisani2004BootstrapEF}.}
    \label{fig:statistical_significance}
\end{figure*}

\noindent \textbf{Training settings.} \canary{} uses FastConformer encoder of XL size from ~\cite{rekesh2023fast}. Along with convolution sub-sampling block, it has 24 conformer layers with model dimension 1024, projection dimension 4096, convolution kernel size 9 and 8 attention heads, with a total parameter count of 0.6B. The decoder is a  regular transformer decoder~\cite{vaswani2017attention}, with 24 layers of dimension 1024, projection dimension 4096, and 8 attention heads, with a total parameter count of 0.4B. The decoder uses fixed-positional encoding.
The encoder consumes audio in the form of 128-dim log-mel features extracted every 10 msec over a window of 25 msec (preliminary experiments didn't show significant difference between 80-dim and 128-dim log-mel features for ASR, but 128-dim features showed consistent improvement for AST).\\
\indent Lhotse~\cite{Zelasko2021Lhotse} was used for dataloading with a batch duration of 360 sec per GPU, \texttt{quadratic\_duration} of 20 sec, \texttt{buffer\_size} of 20000, \texttt{shuffle\_buffer\_size} of 10000 and \texttt{num\_buckets} as 31. To balance multiple languages and datasets, training examples were sampled based on the probability distribution $p_s \sim \left(\frac{n_s}{N}\right)^{\alpha}$, where $s$ represents a stratum (e.g., a language or a dataset), $n_s$ is the number of hours for stratum $s$, $N$ is the total number of hours, and $\alpha$ is the up-sampling factor~\cite{babu2021xls}. We implemented a two-level hierarchical stratification of the training corpus: initially at the language level and subsequently within each language by dataset. The final weight assigned to a dataset is the product of these two probabilities. In both stratification levels, we set $\alpha=0.5$. Non-speech audio has been treated as a separate language for the purposes of sampling.\\
\indent The model was trained in 2 stages using NVIDIA NeMo~\cite{Harper_NeMo_a_toolkit} framework. In stage-1, the model was trained for 150K updates. We used AdamW with a peak learning rate (LR) of 3e-4 and weight decay of 1e-3. The learning rate was warmed up over 2.5K steps and annealed as per Noam scheduling. The encoder was initialized from an XL version of~\cite{NVIDIAMultilingualSTT2023}, whose training data was a subset of Table~\ref{tab:asr_data}. Encoder initialization helped model converge faster and achieve better metrics overall. The decoder was random initialized. Stage-2 was trained for 75K updates with a peak LR of 2e-5, with remaining hyperparameters being same as stage-1. The main difference between both stages is the inclusion of Non-speech dataset from Table~\ref{tab:asr_data} in stage-2 (Note that this 2 stage training is merely practical convenience to allow for faster experimentation wrt to robustness and is not a necessity). In both stages, 128 A100 (80GB) GPUs were used. 

\section{Results}
\label{sec:results}

We evaluate the \canary{} model on speech recognition (ASR) and speech-to-text translation (AST), and show the results in Table~\ref{tab:asr_wer}
 and Table~\ref{tab:ast_bleu} respectively. We use Whisper, OWSM-v3.1, and SeamlessM4T as baselines by using their official checkpoints and re-running the models on the same test sets. All models use beam search decoding with beam width 5.

\begin{table*}[!]
\centering
\caption{Comparing \canary{} with SOTA models across different domains for English ASR~\cite{open-asr-leaderboard}. \canary{} achieves best average WER exhibiting generalizability across domains. }
\label{tab:hf_leaderboard}
\resizebox{0.8\textwidth}{!}{%
\begin{tabular}{l|c|c|c|c|c|c|c|c}
\hline
\multicolumn{1}{c|}{{Model (WER $\downarrow$)}} & {AMI}   & {Earnings22} & {GigaSpeech} & {LS Clean} & {LS Other} & {SPGISpeech} & {Tedlium} & {Avg. WER}  \\ \hline \hline
Whisper-large-v3                    & 16.01          & \textbf{11.3}       & 10.02               & 2.03              & 3.91              & 2.95                & 3.9              & 7.16          \\ \hline
Parakeet-RNNT-1.1B                  & 17.1           & 15.15               & 9.96                & 1.46              & \textbf{2.48}     & 3.11                & 3.92             & 7.60          \\ \hline
Parakeet-TDT-1.1B                   & 15.9           & 14.65               & \textbf{9.55}       & \textbf{1.39}     & 2.62              & 3.42                & 3.56             & 7.30          \\ \hline \hline
\canary{} (1B)                         & \textbf{13.53} & 12.05               & 10.07               & 1.47              & 2.86              & \textbf{2.02}       & \textbf{3.53}    & \textbf{6.50} \\ \hline
\end{tabular}%
}
\end{table*}

\begin{table*}[!]
\centering
\caption{BLEU scores on speech translation (AST) benchmarks. Annotations from the corresponding datasets come with native punctuation and capitalization and are used without further processing/normalization. SeamlessM4T-large-v2 (2.3B) achieves the best overall BLEU scores. \canary{} (1B) matches or outperforms models of comparable size.}
\label{tab:ast_bleu}
\resizebox{\textwidth}{!}{
\begin{tabular}{l|rrr|rrr|rrr|rrr}
\hline
\multicolumn{1}{c|}{\multirow{2}{*}{Model (BLEU $\uparrow$)}} & \multicolumn{3}{c|}{FLEURS (En $\rightarrow$ X)} & \multicolumn{3}{c|}{mExpresso (En $\rightarrow$ X)} & \multicolumn{3}{c|}{FLEURS (X $\rightarrow$ En)} & \multicolumn{3}{c}{COVOST-v2 (X $\rightarrow$ En)} \\ 
\multicolumn{1}{c|}{} & \multicolumn{1}{l|}{En $\rightarrow$ De} & \multicolumn{1}{l|}{En $\rightarrow$ Es} & \multicolumn{1}{l|}{En $\rightarrow$ Fr} & \multicolumn{1}{l|}{En $\rightarrow$ De} & \multicolumn{1}{l|}{En $\rightarrow$ Es} & \multicolumn{1}{l|}{En $\rightarrow$ Fr} & \multicolumn{1}{l|}{De $\rightarrow$ En} & \multicolumn{1}{l|}{Es $\rightarrow$ En} & \multicolumn{1}{l|}{Fr $\rightarrow$ En} & \multicolumn{1}{l|}{De $\rightarrow$ En} & \multicolumn{1}{l|}{Es $\rightarrow$ En} & \multicolumn{1}{l}{Fr $\rightarrow$ En} \\ \hline \hline
OWSM-v3.1 (1.02B) & \multicolumn{1}{r|}{24.37} & \multicolumn{1}{r|}{11.39} & 16.39 & \multicolumn{1}{r|}{19.29} & \multicolumn{1}{r|}{10.98} & 8.59 & \multicolumn{1}{r|}{13.22} & \multicolumn{1}{r|}{9.35} & 12.38 & \multicolumn{1}{r|}{18.05} & \multicolumn{1}{r|}{23.90} & 24.47 \\ \hline
SeamlessM4T-medium (1.2B) & \multicolumn{1}{r|}{28.30} & \multicolumn{1}{r|}{21.05} & 37.36 & \multicolumn{1}{r|}{9.65} & \multicolumn{1}{r|}{16.23} & 8.64 & \multicolumn{1}{r|}{33.39} & \multicolumn{1}{r|}{21.68} & 30.94 & \multicolumn{1}{r|}{35.60} & \multicolumn{1}{r|}{39.18} & 39.27 \\ \hline
SeamlessM4T-large-v2 (2.3B) & \multicolumn{1}{r|}{\textbf{33.17}} & \multicolumn{1}{r|}{\textbf{23.72}} & \textbf{43.05} & \multicolumn{1}{r|}{21.48} & \multicolumn{1}{r|}{34.89} & 26.04 & \multicolumn{1}{r|}{\textbf{37.06}} & \multicolumn{1}{r|}{\textbf{25.41}} & \textbf{33.70} & \multicolumn{1}{r|}{\textbf{39.96}} & \multicolumn{1}{r|}{\textbf{42.91}} & \textbf{42.12} \\ \hline
Whisper-large-v3 (1.5B) & \multicolumn{1}{r|}{N/A} & \multicolumn{1}{r|}{N/A} & N/A & \multicolumn{1}{r|}{N/A} & \multicolumn{1}{r|}{N/A} & N/A & \multicolumn{1}{r|}{33.40} & \multicolumn{1}{r|}{22.70} & 31.02 & \multicolumn{1}{r|}{34.22} & \multicolumn{1}{r|}{39.20} & 35.49 \\ \hline \hline
\canary{} (1B) & \multicolumn{1}{r|}{32.13} & \multicolumn{1}{r|}{22.06} & 39.50 & \multicolumn{1}{r|}{\textbf{24.42}} & \multicolumn{1}{r|}{\textbf{35.76}} & \textbf{27.96} & \multicolumn{1}{r|}{33.70} & \multicolumn{1}{r|}{22.06} & 31.57 & \multicolumn{1}{r|}{37.92} & \multicolumn{1}{r|}{40.79} & 40.58 \\ \hline
\end{tabular}}
\end{table*}

\subsection{Automatic Speech Recognition (ASR)}

We evaluate all models across four languages on MCV-16.1~\cite{commonvoice:2020}, MLS~\cite{pratap2020mls} and VoxPopuli~\cite{wang-etal-2021-voxpopuli} test sets. For the baselines, we input the audios and their corresponding language IDs to the models' inference APIs. For \canary{}, we additionally include the special token \texttt{<|pnc|>} to ensure all  models produce text with punctuation and capitalization.
We then normalize the ground-truth and predictions using the Whisper-Normalizer~\cite{radford2022robust} before calculating the word error rate (WER).
 
Table~\ref{tab:asr_wer} shows that the Canary model achieves the lowest WER in 10 out of 12 test sets across all languages.
On average, \canary{} model achieves 6.20\% WER on English, 6.27\% WER on German, 4.09\% WER on Spanish and 5.39\% WER on French. In comparison, the second best model, SeamlessM4T-large-v2, with twice as many parameters as ours, achieves 5.42\% WER on English, 7.53\% WER on German, 5.24\% WER on Spanish and 6.65\% WER on French. Figure~\ref{fig:statistical_significance} shows that the 95\% confidence intervals do not overlap for most test sets and systems, indicating that the WER improvements observed for \canary{} in Table~\ref{tab:asr_wer} are statistically significant. 

This demonstrates the advantage of the \canary{} model, achieving state-of-the-art multi-lingual ASR performance with fewer parameters and less training data than contemporary models. 

Further, \canary{} achieves the best average WER of 6.5\% across different test sets, highlighting its superior generalization capabilities in English ASR (Table~\ref{tab:hf_leaderboard}).

\subsection{Speech-to-text Translation (AST)}

To evaluate translating English audios to other languages (En $\rightarrow$ X), we use FLEURS~\cite{conneau2023fleurs} and mExpresso~\cite{barrault2023seamless}, whereas FLEURS~\cite{conneau2023fleurs} and CoVoST~\cite{wang2020covost} were used to evaluate translating audio from other languages to English (X $\rightarrow$ En). Annotations from all test sets have punctuation and capitalization and are used without additional processing.

From Table~\ref{tab:ast_bleu}, we notice that SeamlessM4T-large-v2 achieves the highest BLEU scores on all except mExpresso, which is expected given it has the highest number of parameters and the largest size of training data. Meanwhile, \canary{} model outperforms the SeamlessM4T-medium baseline, which shares similar parameter count but trained on more data, on all test sets. Compared with Whisper-large-v3, \canary{} achieves better results on CoVoST-2 and comparable performance on FLEURS. In addition, \canary{} is able to translate English audios into other languages, while Whisper-large-v3 cannot. From these results, we can see that, despite being the smallest model in its class, the \canary{} model achieves competitive performance on speech-to-text translation. 

\vspace{-0.2cm}
\subsection{Long-form ASR Inference}
\vspace{-0.1cm}
We investigate the performance of the \canary{} model on long-form audio by chunking long audios into non-overlapping 30-second segments, performing inference on each segment, and then stitching the transcripts together. We use the FastConformer~\cite{koluguri2023investigating} as our baseline, and show the results on Tedlium3~\cite{hernandez2018ted3}, Earnings21~\cite{del2021earnings21} and Earnings22~\cite{del2022earnings22} in Table~\ref{tab:longform}.  WER's for baselines are copied from the original paper~\cite{koluguri2023investigating}. We can see that, although chunking is a naive method, \canary{} is achieves lowest WER in transcribing long-form audios. Meanwhile, adding streaming capability to \canary{} remains a direction for future research.

\begin{table}[!]
\caption{WER(\%) on long-form ASR inference, both ground-truth and predictions were processed by WhisperNormalizer~\cite{radford2022robust}. All models use greedy decoding. The FastConformer baseline uses a streaming mechanism, while the \canary{} model uses simple chunking without overlap. \canary{} achieves lowest WER.}
\label{tab:longform}
\resizebox{\columnwidth}{!}{
\begin{tabular}{l|r|r|r}
\hline
Model (WER $\downarrow$) & \multicolumn{1}{l|}{Tedlium3} & \multicolumn{1}{l|}{Earnings21} & \multicolumn{1}{l}{Earnings22} \\ \hline \hline
FastConformer-CTC (FT+LCA+GT)~\cite{koluguri2023investigating} & 5.53 & 15.61 & 22.37 \\ \hline
FastConformer-RNNT (FT+LCA+GT)~\cite{koluguri2023investigating} & 4.98 & 13.84 & 19.49 \\ \hline \hline
\canary{} (1B) & \textbf{4.68} & \textbf{11.34} & \textbf{14.34} \\ \hline
\end{tabular}}
\end{table}

\vspace{-0.2cm}
\subsection{Hallucination Robustness}
\vspace{-0.1cm}
The robustness of ASR models is evaluated on many axes, such as robustness to noise, music, background speech, and multiple speakers talking simultaneously. For AED models trained with the next token prediction objective, a particularly less-studied failure case is the generation of spurious transcripts when an audio sample is provided with long periods of silence (or contains no speech). In such cases, the expected output from the model should be an empty transcript. However, autoregressive AED models often hallucinate an unaligned transcript, especially when trained on web-scale data with insufficient filtering. In this work, we investigate the frequency of such hallucinations in the \canary{} model. To that end, we transcribe recordings without any speech. 

Table~\ref{tab:noise} compares the number of hallucinated characters per minute produced by \canary{} with and without noise-robust training (utilizing the strongly-labeled subset of AudioSet from \cite{kim-etal-2019-audiocaps}). We also include Whisper-large-v3 as an additional baseline. As shown in the table, \canary{} generates 16.7\% fewer hallucinated characters than Whisper-large-v3, even without noise-robust training. With noise-robust training, \canary{} further reduces its hallucinated characters by another 26.

\begin{table}[]
\centering
\caption{Number of hallucinated characters per min, measured using 48-hour non-speech audio subset from MUSAN~\cite{musan2015}. \canary{} hallucinates the least. (Note that there is some vocals in MUSAN audios, so the actual number of hallucinated characters may be smaller.)}
\label{tab:noise}
\resizebox{0.8\columnwidth}{!}{
\begin{tabular}{l|r}
\hline
Model & \#  Hallucinated Chars / min ($\downarrow$) \\ \hline \hline
Whisper-large-v3 & 114.8 \\ \hline
\begin{tabular}[c]{@{}l@{}} \canary{} (1B) w/o noise \\ robust training\end{tabular} & 95.61 \\ \hline \hline
\canary{} (1B) & \textbf{70.75} \\ \hline
\end{tabular}}
\end{table}

\section{Conclusions}
\label{sec:conclusions}

In this work, we present \canary{}, a FastConformer-based encoder-decoder ASR and AST model for English, German, Spanish, and French, outperforming similarly sized models on established benchmarks. 
We demonstrate that it is possible to match or exceed the performance of contemporary AST models using solely pseudo-labeled translation data. 
\canary{} was trained using 86K hours of data—an order of magnitude less than contemporary models—while achieving comparable or superior metrics. 
We describe effective training techniques, including encoder initialization, data balancing, and dynamic bucketing batching, enabling us to train the model in under two days. 
The model and code will be open-sourced through NeMo~\cite{Harper_NeMo_a_toolkit}.

\bibliographystyle{IEEEtran}
\bibliography{mybib}

\end{document}